\documentclass{article}

\usepackage{arxiv}

\usepackage[utf8]{inputenc} 
\usepackage[T1]{fontenc}    
\usepackage{url}            
\usepackage{booktabs}       
\usepackage{amsfonts}       
\usepackage{nicefrac}       
\usepackage{microtype}      
\usepackage{lipsum}
\usepackage{bm}
\usepackage{graphicx}
\usepackage{amsmath}
\usepackage{float}
\graphicspath{ {./images/} }
\usepackage{caption}
\usepackage{subcaption}
\usepackage{multirow}
\usepackage[normalem]{ulem}
\useunder{\uline}{\ul}{}
\usepackage{enumitem}
\usepackage{authblk}
\usepackage{url}
\usepackage{setspace}
\usepackage[hidelinks]{hyperref}
\usepackage{natbib}
\usepackage{pdflscape}
\usepackage{rotating}

\usepackage{algorithm,algpseudocode}
\usepackage{amsmath}
\usepackage{amsfonts}
\usepackage{amssymb}

\usepackage{ifthen}

\usepackage{xcolor}

\usepackage[compact]{titlesec}
\titlespacing{\section}{0pt}{2ex}{1ex}
\titlespacing{\subsection}{0pt}{1ex}{0ex}
\titlespacing{\subsubsection}{0pt}{0.5ex}{0ex}

\begin{document}
\title{A Supervised Machine Learning Model For Imputing Missing Boarding Stops In Smart Card Data}

\author[1]{Nadav Shalit}
\author[1]{Michael Fire}
\author[2]{Eran Ben-Elia} 
\affil[1]{Department of Software and Information Systems Engineering, Ben-Gurion University, Israel }
\affil[2]{GAMESLab, Department of Geography and Environmental Development, Ben-Gurion University, Israel}
\affil[]{\textit {\{shanad,mickyfi,benelia\}@bgu.ac.il}}

\maketitle
\begin{abstract}
Public transport has become an essential part of urban existence with increased population densities and environmental awareness. Large quantities of data are currently generated, allowing for more robust methods to understand travel behavior by harvesting smart card usage. However, public transport datasets suffer from data integrity problems; boarding stop information may be missing due to imperfect acquirement processes or inadequate reporting. We developed a supervised machine learning method to impute missing boarding stops based on ordinal classification using GTFS timetable, smart card, and geospatial datasets. A new metric, Pareto Accuracy, is suggested to evaluate algorithms where classes have an ordinal nature. Results are based on a case study in the city of Beer Sheva, Israel, consisting of one month of smart card data. We show that our proposed method is robust to irregular travelers and significantly outperforms well-known imputation methods without the need to mine any additional datasets. Validation of data from another Israeli city using transfer learning shows the presented model is general and context-free. The implications for transportation planning and travel behavior research are further discussed.
\end{abstract}

\section{Introduction}\label{Introduction}
Public transport (PT) is an integral part of everyday life in many cities. The gradual shift of the global population over the past century to urban areas is markedly increasing people's dependence on PT for their daily mobility needs \citep{petrovic2016appraisal}. PT is a complex system that is based on physical elements of stops, vehicles, routes, and other temporal and spatial elements \citep{ceder2016public}. The PT system consists of regularly scheduled vehicle trips open to all paying passengers, with the capacity to carry multiple passengers whose trips may have different origins, destinations, and purposes \citep{walker2012human}. PT is ideal when passengers regard its service as punctual and regular \citep{walker2012human}. With the growth in the number of cars on urban roads, PT improvements have become an essential part of traffic congestion mitigation strategies and vital in promoting sustainable transportation \citep{al2011composite}.
Although understanding the patterns of PT use is crucial to its planning, this task remains a significant challenge in practice and research.

\begin{landscape}
\begin{figure}
    
\centering
    \includegraphics[scale=0.75]{"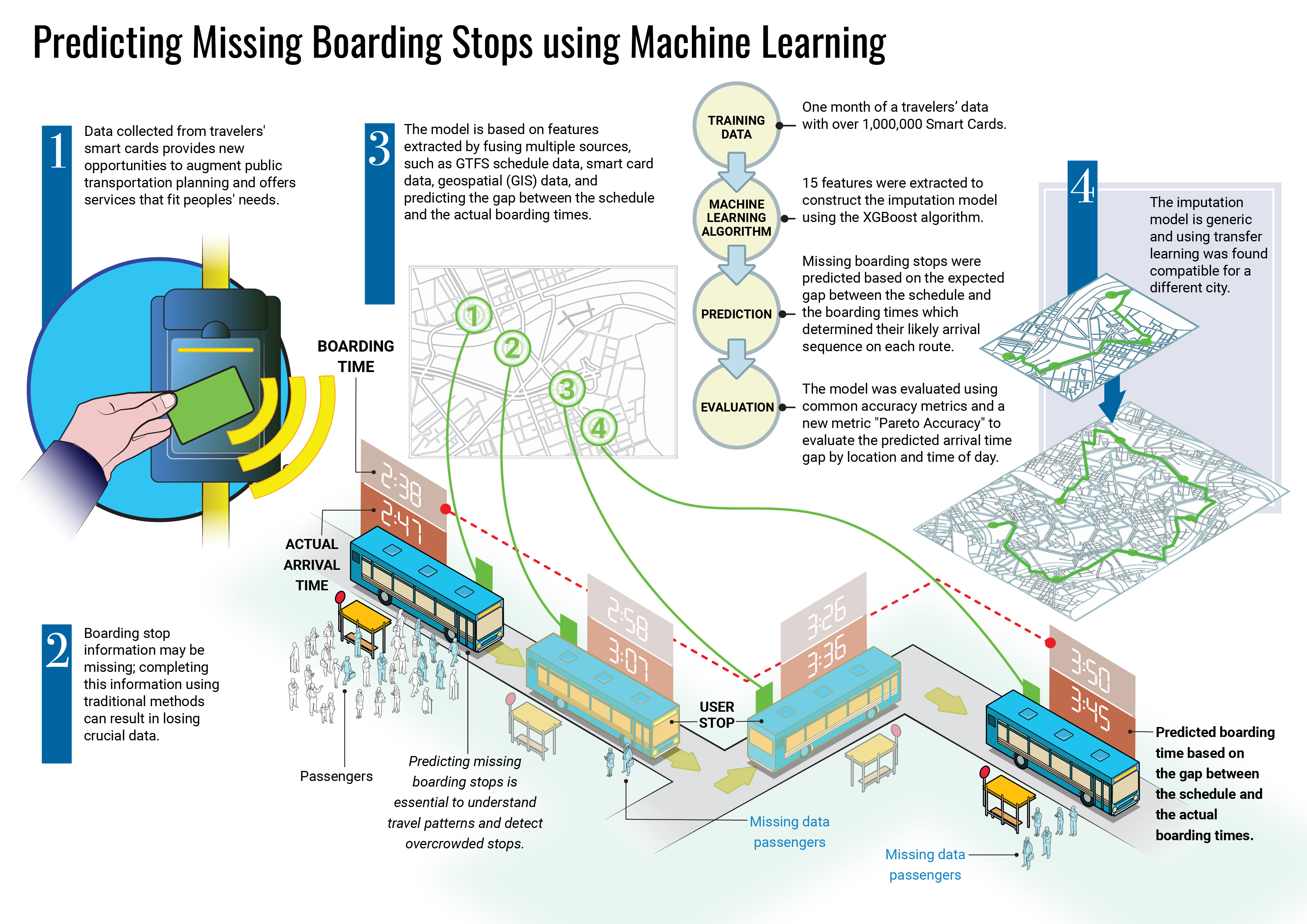"}
    \caption{Predicting missing boarding stops algorithm overview.}
    \label{fig:infographic}
\end{figure}
\end{landscape}

Numerous studies in recent years have examined the behavior of PT travelers \citep{li2018smart} in efforts to address this challenge. Habitual travel behavior is of great interest to transportation planners, and its analysis can help improve demand predictions and justify necessary upgrades to PT supply \citep{briand2017analyzing}. This analysis can also contribute to improvements in PT (service/planning/upgrades) with respect to the management of COVID transmission, in terms of providing better information on crowded areas, such as bus stops, which is critically important to the global issue of public health. To this end, transportation planners typically use travel behavior surveys \citep{stopher2007household}. While these surveys statistically reflect travel behavior correctly, they are also expensive, time-consuming, and often unable to generate sufficient amounts of data relative to the size of the population, , and would need significant changes in scope to cover recent COVID concerns.

Conversely, data harvested from smart cards can generate millions of records compared to a typical sample ranging from 2,500 to 10,000 households using surveys  \citep{maeda2019detecting}.
Smart cards, also known as automatic fare collection (AFC), provide an efficient and cost-saving alternative to the manual fare collection method
 \citep{jang2010travel,chen2018extracting}. In addition to fulfilling fare collection needs, as a bi-product, smart card transactions also generate geocoded timestamps that record every passenger's boardings, line transfers, and sometimes alightings for a wide range of PT vehicles (bus, tram, train, or metro) \citep{faroqi2018applications, pelletier2011smart}. These records are generated for almost the entire passenger population~\citep{faroqi2018applications, pelletier2011smart}. Such information is a treasure trove for travel behavior analyses, especially for extracting passengers' spatiotemporal travel patterns (e.g., Origin-Destination matrices or path choice \citep{wang2011bus}). Nevertheless,  common statistical inference methods applied in surveys are of little practical use for understanding the travel patterns of an entire population. Therefore, different methods are required. 
 
Kandt and Batty (\citeyear{kandt2020smart}) proclaimed a new area of urban research defined by advances in big data analytics, with smoother decision-making and a deeper understanding of urban systems. A massive increase in the \emph{volumes}, \emph{velocities} and \emph{varieties} of big data have also been paralleled by recent developments in the data science field. New data mining tools and robust cloud computing capabilities \citep{li2018smart, li2015towards} create new opportunities to analyze travel behavior patterns at the individual level, over extended periods, and in large urban areas \citep{ma2017understanding}. The availability of big data has a vast potential to improve the quality of transportation planning and research, and by applying big data analytics and data mining methods, this task has become much more feasible \citep{ma2013mining}. 

However, similar to the case in other domains, the \emph{veracity} of such datasets remains questionable \citep{ben2018epilogue}. Smart card datasets, in particular, may suffer from integrity problems, such as incorrect or missing values, e.g., when operators only record partial data. For example, in Yan et al. (\citeyear{yan2019alighting}), boarding stop information was completely missing and only time stamps remained intact in the dataset. A common solution for such problems is to replace the missing or erroneous data by utilizing alternative publicly accessible data. One possible solution is to use official PT timetables to impute the missing data in missing boarding stop information. One popular source for such data comes from the General Transit Feed Specification (GTFS), first created in 2006 by Google (Google, 2016), defined as a standard file format for storing PT schedules and associated geographic information \citep{ma2012transit}. GTFS contains the complete schedules and routes of every PT line planned for each day of the month in tabular formats together with corresponding geographic shapefiles and is now widely used in over 750 urban regions across the world \citep{hadas2013assessing, antrim2013many}. 

Nonetheless, PT running times and arrival times at stops are never perfectly aligned with their official timetables, where PT is not always punctual, even in developed countries. For example, Cats and Loutos (\citeyear{cats2016real}) found that only 10\% of all arrivals were within an interval of 15 seconds. This issue becomes more acute, especially when PT vehicles - mainly buses - also share the same road space with private and commercial vehicles (i.e., mixed traffic). While this issue is less severe in major urban areas in developed countries where rail-based and PT bus preemption infrastructure is widespread and right-of-way strongly enforced, this is not the reality everywhere. For example, in Israel (an official OECD member), buses accounted nationally for 85\% of PT trips in 2019, with more than 2M passengers served daily. The country suffers from a shortage in adequate PT infrastructure (namely too few priority lanes - 14 meters per capita, compared to 300 meters in the EU), thus resulting in poor PT service punctuality \citep{ceder2004new}. As shown, later on, this fact makes schedule-based imputation a poor substitute for boarding stop prediction.
A second solution is to discard such data by simply removing missing records or those that do not align with a prescribed hypothesis \citep{tao2014examining}. Nonetheless, discarding data can be regarded as a reasonable solution only when that share of the missing data is small. However, when the missing portion is substantial, the whole dataset could be compromised and discarded. This scenario can render certain urban areas effectively blind vis-a-vis smart card data. A third option is to complement the missing data by combining different datasets. In this respect, either automatic vehicle location (or AVL) which uses installed GPS  transponders to locate PT vehicles and estimate real-time arrival times at designated stops; or automatic passenger counters (or APC) which use infrared or laser technologies to estimate boarding and alighting passenger numbers, have been used in combination with smart card data \citep{mazloumi2010using,shalaby2004prediction,khiari2016automated}. 
Yet, such data is neither always available \citep{yan2019alighting, chen2018extracting}, nor efficient, as many more errors may well be introduced in the process \citep{luo2018constructing}. These two facts likely reduce their suitability for data imputation. Moreover, even when such data sources exist, matching between them is somewhat challenging. For example, Lou et al. (\citeyear{luo2018constructing}) had no vehicle trip identification (ID), making it impossible to match with AVL records. A further difficulty is that missing data can vary by city or between different operators \citep{lana2018imputation}. In some cities, data integrity is regarded as very strong, and consequently, boarding and alighting imputation tasks are very good \citep{munizaga2014validating}. In contrast, in other cities where data sources are lacking, data integrity can also be flawed. Therefore heuristic methods, such as ML, are the only viable solution to perform imputation tasks  \citep{yan2019alighting}.

To this end, our aim is a general and context-free boarding stop imputation method. Specifically, we address use cases where data quality is considered too insufficient to impute by cross-inference and without the need to harvest any other data than what is necessary. While still providing valuable insights for transportation planners, we consider this of particular relevance to developing countries where the traveler population is mostly PT-dependent. We established a general boarding stop imputation method to improve the quality and integrity of PT datasets by predicting missing or corrupted travelers’ records in smart card data. 

Namely, to the best of our knowledge, we developed the first machine learning (ML) algorithm for predicting passengers’ boarding stops (see Figure~\ref{fig:infographic}). Our algorithm is based on features extracted by harvesting three big data sources, the planned GTFS schedule data, smart card (AFC) data, and geospatial (GIS) data. We applied a machine learning model to these features to predict boarding stops based on the notion of embedding (see Section \ref{Methods}). To train and evaluate our algorithm’s performance, we utilized a real-world smart card dataset from the city of Beer Sheva in Israel that consists of over a million trips taken by more than 85,000 passengers. Since the boarding stops are embedded, they also become ordered, and therefore, the problem we addressed is ordinal classification. Accordingly, we also propose a new method of evaluation that shows the percentage in each error dimension that we define as Pareto Accuracy, which is more interpretable and allows for better comparison between imputation models. We show that our model performed significantly better than a naïve prediction model based on harvesting GTFS data alone (aka schedule-based) and other imputation methods.

In this study, we succeeded to generate a model which is both wholly generic and has considerably higher accuracy and recall values than other tested imputation methods (see Section \ref{Results}). Additionally, we demonstrated that we obtain similar prediction results in an entirely different city using our method. Moreover, we show how other imputation methods are not always applicable, while our methodology can be applied with broader scope.

\subsection{Contributions}\label{Contributions}
Our study’s overall focus is to improve the integrity of public transport data.  Specifically, our study provides the following two main contributions:
\begin{enumerate}
  \item We present a novel prediction model for imputing missing boarding stops using supervised learning. Moreover, our proposed model is generic and transferable, i.e., it can be trained on one city's data and then impute missing data in another municipality.
  \item We developed a new method - Pareto Accuracy -  for evaluating public transport metrics that are more interpretable and allow broader comparisons between imputation models.
\end{enumerate}

\subsection{Organization}\label{Organization}
The rest of the paper is organized as follows: In Section \ref{Related Work}, we review related work on smart card usage, missing data imputation, and ML applications in transportation research. Section \ref{Methods} describes the use case, experimental framework, and methods used to develop the ML model and the extraction of its features. In Section \ref{Results}, we present the results of the ML model and compare its performance to other known solutions. In Section \ref{Disccusion}, we discuss the implications of the findings and the study's limitations and present our conclusions and future research directions.

\section{Related work}\label{Related Work}
We provide an overview of relevant studies by first presenting smart card research in general, followed by studies that have utilized smart card data with machine learning to perform predictive analytics. We then give an overview of the field of missing data imputation. Lastly, we present studies in the field of ordinal classification.

\subsection{Smart card analytics}\label{Smart Card}
The smart card system was introduced as a smart and efficient AFC system in the early 2000s \citep{chien2002efficient} and has since become an increasingly popular payment method \citep{bagchi2005potential,trepanier2007individual}. In particular, smart cards have also become an increasingly popular source of big data for research and policy making  \citep{jang2010travel,agard2007mining}. For example, smart data is used for exploring travel behavior, determining travel patterns, measuring the performance of PT services, locating critical transfer points, and analyzing crowdedness effects on route choice \citep{bryan2007understanding, alguero2013using, zhao2017spatio, li2018smart,jang2010travel, yap2020crowding}. Recently, smart card datasets were used to study travel behavior changes to travel behavior as a result of the Covid-19 pandemic~\citep{zhang2021changes,almlof2020still,orro2020impact}.  Comprehensive literature reviews of smart card usage were provided by  \citep{pelletier2011smart},  \citep{faroqi2018applications} and \citep{schmocker2017overview}.

Initially, smart card research applied rather classic statistical methods and descriptive analytics. Devillaine et al. (\citeyear{devillaine2012detection}) inferred location, time, duration, and designation of PT users’ activities using rules derived from smart card data and work and study schedules. The main research challenge evident in the literature was to estimate origin-destination (OD) matrices which describe the spatial distribution of travel demand between locations during different periods of the day \citep{wang2011bus, gordon2013automated, munizaga2012estimation, chu2008enriching}. OD matrices are also crucial inputs to perform the three stages in PT network design, namely: route design, frequency (headway) setting, and timetabling \citep{guihaire2008transit}. Before the advent of smart cards, these matrices were only derived and validated based on some representative sample of travelers \citep{chen2016promises}. However, as noted, surveys often lack sufficient spatial and temporal coverage. Various studies have demonstrated the advances in OD estimation with smart card data \citep{wang2011bus, gordon2013automated, munizaga2012estimation, chu2008enriching}. 

Nevertheless, with the introduction of smart cards, new problems in OD estimation appeared. Namely, many PT agencies adopted a TAP (Transit Access Protocol) IN system where only boarding stop information is recorded. In contrast, the availability of alighting stop information  ``TAP IN+TAP OUT'' systems allows for the OD matrix to be derived using more straightforward approaches. Alighting stop information is necessary for many tasks such as route loading profiles, market research, and improvements in service planning. However, under TAP IN,  the destination must be somehow predicted \citep{faroqi2018applications, trepanier2007individual}.

 In addition, combining smart card data with a smaller scale travel behavior survey for validation purposes is a useful approach to better understand passengers' daily travel patterns \citep{wang2011bus}. Nonetheless, OD analyses inherently assume that PT passengers travel routinely back and forth from/to the same locations. Recent findings suggest this assumption does not necessarily hold, and some share of PT passengers are quite flexible \citep{huang2018tracking} or use PT infrequently \citep{benensonservicing}. Therefore, a simple OD estimation will possibly result in PT planning that is mismatched with actual demand patterns.

Traditional analysis methods do not take advantage of the full potential of the added value of big data. At the same time, rapid growth in power and cost reduction in computational technologies provide new opportunities, both in terms of the availability of the massive amount of data collected and the development of more novel algorithms \citep{welch2019big}. Agard et al. (\citeyear{agard2007mining}) obtained travel behavior indicators that identify daily travel patterns and clustering of major user groups. Bhaskar et al. (\citeyear{bhaskar2014passenger}) applied a density-based spatial clustering application with noise (DBSCAN) algorithm to cluster passengers and identify classes of passengers for strategic planning improvements. Ma et al. (\citeyear{ma2013mining}) used smart card data to cluster the travel patterns of PT riders to characterize commuter profiles.

In this respect, the literature shows a shift toward harvesting the prognostic nature of machine learning (ML) to yield better predictive analytics highlighting the growing emphasis on using smart card data for analytical purposes. This shift underscores the change from the more straightforward analyses conducted in the past to the more comprehensive analysis done today. Hagenauer and Helbich (\citeyear{hagenauer2017comparative}) compared several ML classifiers and showed both their predictive power and ability to uncover travelers' mode choices via feature importance analysis. For example, the trip distance was the most important predicting factor, while the temperature was only a key feature for predicting bicycle use (\citeyear{hagenauer2017comparative}).  In 2018, Palacio (\citeyear{palacio2018machine}) showed that ML predictions are much more accurate than traditional linear models that were sub-optimal both in terms of R-square and MSE. In the following year, Traut and Steinfeld (\citeyear{traut2019identifying}) combined smart card data with crime records to assist agencies in identifying insecure, and dangerous PT stops. Chen et al. (\citeyear{chen2016promises}), who inferred mode and route choices, stress the need for cross-disciplinary collaborations between data scientists and transportation planners to exploit the information withheld in the data. Further evidence of the prominence of big data analytics in PT research can be found in several review papers such as \citep{fonzone2016new, anda2017transport, milne2019big, li2018smart, namiot2017survey}.

Deep learning algorithms have also been utilized to address PT issues using smart card data. Deep learning is a sub-field of ML that automatically creates feature engineering, and its methods are state-of-the-art in many domains.
Examples of such implementations include inference of passenger employment status \citep{zhang2019deep}, forecasting passenger destinations \citep{jung2017deep,toque2016forecasting} using standard deep network and long- and short-term memory networks, inference of demographics using convolutional neural networks \citep{zhang2019deep2}, improving passenger segmentation \citep{chen2018traveler}, and predicting multimodal passenger flows \citep{toque2017short}.

\subsection{Missing data imputation}\label{Missing Data Imputation}
Incomplete data is a universal problem, and the application of different imputation methods will often yield different results. Therefore, to preserve reproducibility, they must be adequately addressed \citep{saunders2006imputing}. This problem is, notably, relevant for transportation planning, e.g., in the case of road traffic analysis \citep{qu2009ppca}. Incomplete data is a well-known problem in the data mining literature where a significant amount of data can be missing or incorrect. Lakshminarayan et al. (\citeyear{lakshminarayan1996imputation}) elucidated both the severity of this issue as well as recommended applying ML techniques toward its solution rather than classical statistical methods. Batista and Monard (\citeyear{batista2003analysis}) assert that missing data imputation must be carefully handled to prevent bias from being introduced.
Moreover, they show that the most common methods, such as mean or mode imputation, are not always optimal. One example we found in the PT literature is of Kusakabe and Asakura (\citeyear{kusakabe2014behavioural}). They used a Naïve Bayesian model for data imputation and analysis of PT to understand continuous long-term changes in trip attributes. They showed both the power of smart card data and the usefulness of missing data imputation in this field. Their method of imputation, however, is not reported in sufficient detail to be understood or replicated. 

Several techniques to optimize missing data imputation showed the importance attributed to this area of research  \citep{bertsimas2017predictive}. Moreover, even state-of-the-art deep learning methods have been applied to this problem  \citep{camino2019improving, garg2018dl, costa2018missing}. These implementations were performed on a variety of datasets and problems such as classification of continuous attributes (breast cancer and default credit card classification); images \citep{camino2019improving,garg2018dl} and regression \citep{camino2019improving}. Insofar as this field of study has not been operationalized for PT data,  further examination is warranted, particularly when considering the issue of completing missing data to provide better information on crowded PT areas, as it pertains to the spread of COVID.

In many imputation tasks, including PT, ML methods outperform standard methods significantly when the missing portion increases \citep{yan2019alighting, saunders2006imputing, lana2018imputation, echaniz2019modelling}. Additionally, standard imputation methods are too sensitive to the ratio of missing data and infrequent or ‘irregular’ users of the PT network \citep{van2005accurate}. Conversely, ML-based imputation showed stable results regardless of the missing ratio \citep{lana2018imputation}. As noted previously, one solution is to impute the missing boarding stops using complementary datasets such as AVL or APC. However, AVL data are not always available \citep{chen2018extracting} whereas combining several datasets (i.e., AVL, AFC, APC, GTFS, etc.) can introduce more errors, and make it much harder to match them perfectly \citep{luo2018constructing}.

\subsection{Ordinal classification}\label{Ordinal classification}
Classification is a form of supervised ML that aims to generalize a hypothesis from a given set of records. It learns to create $h(x_i)\rightarrow y_i$ where $y$ has a finite number of classes \citep{kotsiantis2007supervised}. The basic metrics for classification are sensitivity, specificity, and accuracy \citep{jiao2016performance}. Accuracy is the percentage of observations classified correctly, specificity is the percentage of true negatives classified correctly, and sensitivity is the percentage of true positives classified correctly.
 A classification task becomes ordered when the classes have some inherent order between them. However, the aforementioned metrics (e.g., sensitivity, specificity, and accuracy) for evaluating this problem are unsuitable, in our case, as they also require a high level of interpretability of the results.
 
Ordinal classification is a form of multi-class classification where the classes exhibit some natural ordering (such as cold, warm, and hot), but not necessarily numerical traits for each class. Rather than being chosen based on the traditional metrics discussed above, a classifier may be chosen based on the severity of its errors \citep{gaudette2009evaluation}. Additionally, classic modeling techniques will sometimes perform suboptimally since ML models assume there is no order between classes. In such tasks, e.g., the well-known Boston housing and breast cancer datasets, different models that take advantage of ordinal information are preferred \citep{frank2001simple}. In this case, additional metrics are proposed to calculate such tasks differently, such as regression metrics like Mean Absolute Error (MAE) and the Mean Square Error (MSE) and even their own metric, the Ordinal Classification Index \citep{cardoso2011measuring}. Notwithstanding, as noted below, these approaches neither fit our data nor our needs. Therefore, we developed a different and novel performance metric (see Section~\ref{Model Evaluation}).

\section{Methods}\label{Methods}

The main goal of our study is to use ML algorithms to improve the integrity of PT data. Specifically, we develop a supervised learning-based model to impute missing boarding stops in any given smart card dataset. Moreover, our goal is to construct a generic model that will be fully transferable to other datasets to impute missing data in different contexts without further adjustments.

To maintain these generic objectives, we had to contend with two significant challenges:  First, we could only incorporate generic properties in our model. For instance, our model cannot include the actual line number of a bus route specific to a particular city. Moreover, since supervised ML algorithms can only predict classes they were initially trained upon, classification classes must remain the same across datasets, e.g., bus stop \#14 in a specific city is an irrelevant feature for other cities. Therefore, once more, a different numerical representation is applied by embedding (see Section \ref{Feature extraction and machine learning model construction}). Second, in order to develop a genuinely generic model that can also be applied to other geographical contexts on which it was not initially trained, the model must also undergo a process of transfer learning \citep{torrey2010transfer} that entails the transfer of relevant knowledge by fine-tuning a model on a "novel" dataset. In our case,  our model underwent the process of transfer learning using a dataset on which it had not trained before. 
  
The missing boarding stop values were imputed using the following methodology (see Figure \ref{fig:infographic}): First, we preprocessed and cleaned the smart card dataset that we utilized in this study (see Section \ref{Datasets and data preprocessing}). Next, we extracted various features from two other datasets: (a) the GTFS timetable data; and (b) open municipal geospatial data. In addition, we converted boarding stops from their original identifiers to embedded numerical representations based on GTFS data (see Section \ref{Feature extraction and machine learning model construction}). Afterward, we applied ML algorithms to estimate a model that can predict the missing boarding stops. We used SHAP values for determining feature importance \citep{lundberg2017unified}, i.e., which features make the most substantial contribution to the predictive power of the model (see Figure \ref{fig:Feature_importance}). We also evaluated the performance of our model using a novel metric called Pareto Accuracy. Then, based on common metrics, we evaluated our model relative to a schedule-based model estimated only on GTFS timetable data. Finally, we compared our model to several other comparative models (e.g., passenger history, temporal proximity, or semi-random guessing) that were previously used in the literature. Below, we describe each step of our approach in more detail.

\subsection{Datasets and data preprocessing}\label{Datasets and data preprocessing}
As noted above, we used three datasets: 
 \begin{enumerate}[topsep=1pt, partopsep=1pt,noitemsep]
 
  \item \textit{The Smart card dataset} - ``Rav Kav'' is the Israeli AFC system applying the TAP protocol, allowing PT passengers to pay for their trip using their smartcards anywhere in the country. Rav-Kav operates a nationwide TAP IN for buses and rail that codes information on unique passenger identifiers, traveler types (such as student or senior travelers), boarding stops, boarding timestamps, fares, discount attributes, and unique trip identifiers of the line at that time. For rail trips only, TAP OUT also records alighting stops and times. During the period 2018/9, circa 2M boardings were recorded per day in the entire country. 
   
  \item \textit{GTFS} - a GTFS feed, as described above, consists of rail/bus schedules and timetables, stops, and routes of every PT trip planned for every day of the month. In Israel since 2012, the GTFS feed is published daily online by the Ministry of Transport, providing schedules of 36 bus and rail operators, encompassing 7,800 route-direction-pattern alternatives served by 28,000 bus and rail stations. The GTFS feed aligns with the smart card dataset as described below. This study utilized the GTFS dataset to enrich the feature space and convert boarding stop records into an embedded numerical value.
  \item \textit{Geospatial information} - we derived a variety of geospatial attributes from municipal GIS databases.
\end{enumerate}
To obtain a dataset suitable for constructing the prediction model, we were required to remove any record that lacked a boarding stop or a trip ID (a unique identifier of a trip provided by a specific and unique PT operator) from the smart card dataset. Next, we joined the smart card dataset with the GTFS dataset by matching the trip ID attributes. Lastly, we joined the geospatial dataset with the smart card dataset using the GTFS dataset, which contains all the geographic coordinates of each PT route.

\subsection{Feature extraction and machine learning model construction}
\label{Feature extraction and machine learning model construction}
ML performance is highly correlated to the quality of the feature space, and therefore, including more features results in better model performance  \citep{gudivada2017data}. While the smart card data contains the PT line and boarding time of each passenger, it lacked several essential data, such as the duration that had elapsed since that line left the origin depot, the time remained until arrival to the final destination, the total number of stops, and other relevant trip attributes. Moreover, the smart card data is missing physical geospatial characteristics, such as the number of traffic lights on the PT route that more likely increases traffic congestion and consequent delays, and could well strengthen model performance.

Overall, three features were extracted using the smart card dataset, five features using the GTFS dataset, three features using the geospatial dataset and four from combined GTFS and smart card datasets. From the 41 features we tested in total, we have selected 15 to include in our model (see Table \ref{tab:features}).

\begin{table}[H]
\caption{\label{tab:features}Extracted Features}
\begin{tabular}{|l|l|l|}
\hline
\textbf{Dataset }&
 \textbf{ Feature} &
  \textbf{Explanation} \\ \hline
\multirow{3}{*}{\begin{tabular}[c]{@{}l@{}}Municipal \\   geospatial\\ records\end{tabular}} &
  Addresses\_average &
  \begin{tabular}[c]{@{}l@{}} The number of addresses listed on the route\end{tabular} \\ \cline{2-3} 
 &
  Street\_light\_average &
  \begin{tabular}[c]{@{}l@{}} The number  of streetlights on the route\end{tabular} \\ \cline{2-3} 
 &
  Traffic\_Lights\_average &
  \begin{tabular}[c]{@{}l@{}} The number of traffic lights on the route\end{tabular} \\ \hline
\multirow{5}{*}{GTFS} &
  Number\_of\_points &
  \begin{tabular}[c]{@{}l@{}} The number of points in a shapefile in\\  GTFS per route\end{tabular} \\ \cline{2-3} 
 &
  Average\_distance\_per\_stop &
  \begin{tabular}[c]{@{}l@{}}The total length of the route divided by\\ the number of points\end{tabular} \\ \cline{2-3} 
 &
  Average\_time\_per\_stop &
  \begin{tabular}[c]{@{}l@{}}The total expected travel time of the route  \\divided by the number of points\end{tabular} \\ \cline{2-3} 
 &
  Average\_points\_to\_stops &
  \begin{tabular}[c]{@{}l@{}} The number of points in a shapefile in \\ GTFS per route divided by the number of points\end{tabular} \\ \cline{2-3} 
 &
  Time\_diff\_of\_trip &
  Total travel time \\ \hline
\multirow{4}{*}{\begin{tabular}[c]{@{}l@{}}GTFS and\\   smart card\end{tabular}} &
  Time\_from\_boarding\_to\_last\_stop &
  \begin{tabular}[c]{@{}l@{}}Time from boarding time to expected last stop of the route\end{tabular} \\ \cline{2-3} 
 &
  Time\_from\_departure\_to\_boarding &
  Time from route departure time to boarding time \\ \cline{2-3} 
 &
  Predicted\_sequence &
  GTFS prediction sequence of the most likely stop \\ \cline{2-3} 
 &
  Hourly\_expected\_lateness &
  The average lateness per hour (based on training data) \\ \hline
\multirow{3}{*}{\begin{tabular}[c]{@{}l@{}} Smart \\ card\end{tabular}} &
  Boardingtime\_Seconds\_from\_midnight &
  \begin{tabular}[c]{@{}l@{}}Timestamp of  boarding to a numerical value in seconds \\ from midnight\end{tabular} \\ \cline{2-3} 
 &
  Boardingtime\_weekday &
  The day of the week in which the boarding occurred \\ \cline{2-3} 
 &
  Is\_weekend &
  Is it a weekend? \\ \hline
\end{tabular}
\end{table}

  To construct the prediction model, we used the GTFS dataset to create a schedule-based prediction. This naive prediction reflects the transit vehicle's position along a line according to the GTFS schedule. Namely, let $S_i$ be the sequence number of the boarding stop based on the GTFS schedule and let $A_i$ be the actual boarding stop sequence number. Then, we define $D_i$ as $D_i= A_i-S_i$. Our prediction model goal was to predict $D_i$ by utilizing the variety of features presented in the previous section.

For instance, consider a passenger who boarded a line at the third stop, i.e., $A_i = 3$, but the transit vehicle was scheduled to arrive at the second stop at the designated time. The schedule-based prediction would be 2, i.e., $S_i = 2$, the stop where it was supposed to be at that time. Then, the difference is $D_i = A_i-S_i = 3-2 = 1$, and this is the class the algorithm will predict.

\begin{figure}[H]
\begin{subfigure}{.5\textwidth}
  \centering
  \includegraphics[height=1.2in]{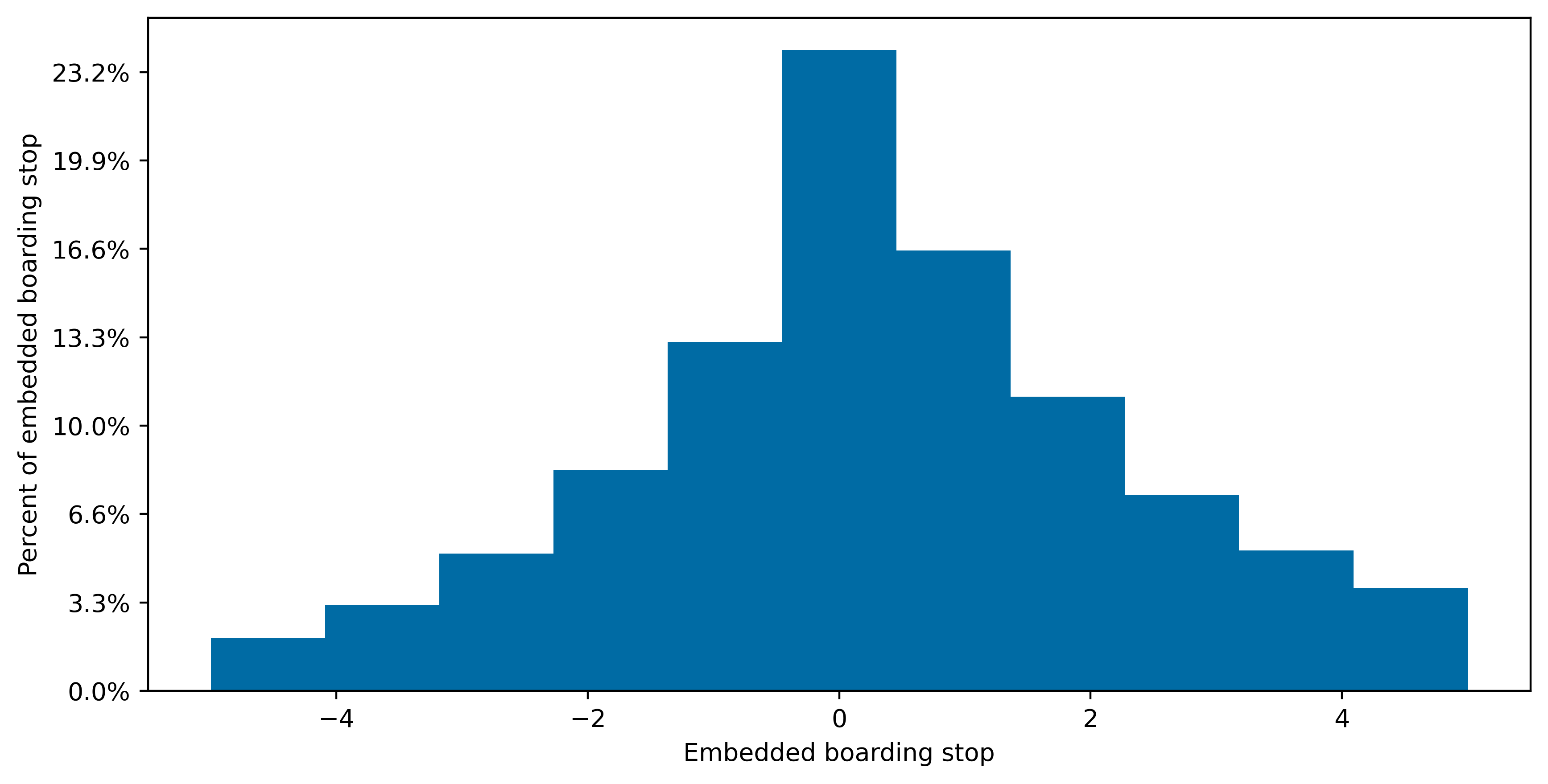}  
  \caption{Train boarding stop histogram}
\end{subfigure}
\begin{subfigure}{.5\textwidth}
  \centering
  \includegraphics[height=1.2in]{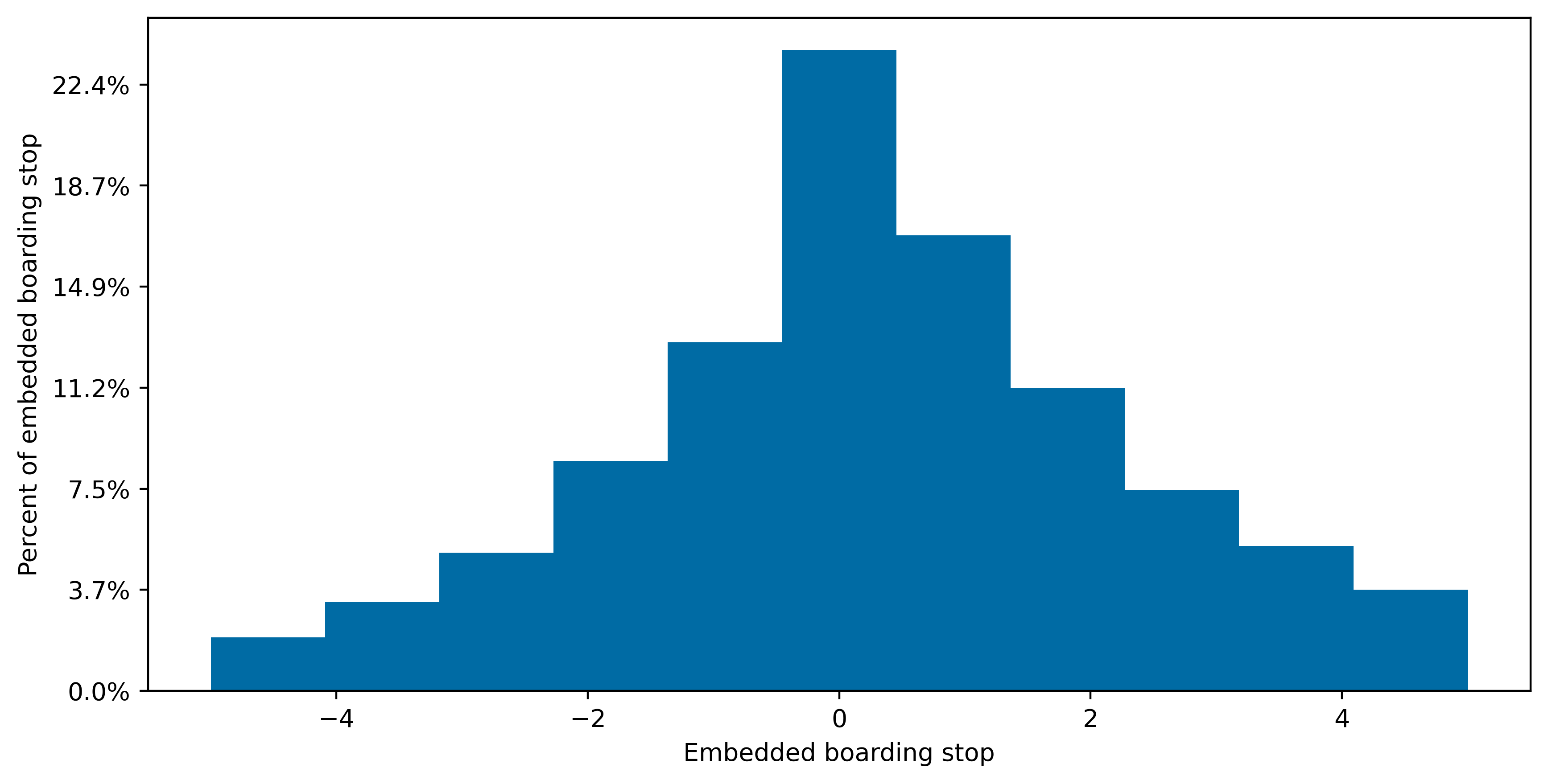}  
  \caption{Test boarding stop histogram}
\end{subfigure}
  \caption{Train and Test boarding stop histogram}
 \label{fig:boarding_stop_histograms}
\end{figure}

Subsequently, we performed the following steps to construct the prediction model: First, we selected several well-known classification algorithms. Namely, we used Random Forest  \citep{singh2016review}, Logistic Regression \citep{singh2016review}, and XGBoost \citep{chen2016xgboost}. Second, we split our dataset into a train dataset, which consisted of the first three weeks of data, and a test dataset, which consisted of the last week of the same dataset. Figure \ref{fig:boarding_stop_histograms} shows the distributions of the embedded boarding stops by the computed difference between the actual and schedule-based sequences ($D_i$) for the train and test subsets. No apparent differences between the two distributions are evident. Third, for both the train and test datasets, we extracted all the 15 features mentioned above. Fourth, we constructed the prediction models using each one of the selected algorithms. Lastly, we compared the generated models and selected the one with the best performance based on the \emph{Pareto Accuracy}  metric (see Section \ref{Model Evaluation}).

\subsection{Model Evaluation}\label{Model Evaluation}
We evaluated each model and compared it to the schedule-based method on the test dataset using common metrics:
\textit{accuracy}, \textit{recall}, \textit{precision}, \textit{F1} (see Appendix A for definitions), and the new metric we developed \textit{Pareto Accuracy}. 
We used the following variables for our novel \textit{Pareto Accuracy} metric: Let $p_i$ be the predicted sequence of ${\rm stop}_i$, $a_i$ be the actual sequence, and $d_i$ be the absolute difference between them. Let $l$ be the limit of acceptable difference for imputation, i.e., if an error of one-stop is tolerated, such as for neighborhood segmentation, then $l = 0$. Let $X_i$ be an indicator defined as: \[X_i=\begin{cases}  1 & \text{if $di <=l$} \\  0 & \text{otherwise}\end{cases}.\]
We defined Pareto Accuracy as follows: \[{\rm PA}_l=\ \frac{\sum_{i=1}^{n}X_i}{n}.\]

The PA metric is a generalization of the accuracy metric. Namely, $PA_{0}$ is the well-known accuracy metric.
Unlike other ordinal classification methods, the primary advantage of using the PA metric is to evaluate the accurate dimension of error while being extremely robust to outliers (by setting parameter $l$). Moreover, this metric is highly informative since its outcome value can be interpreted easily; for example, 0.6 means that 60\% of predictions had at most $l$ difference from true labels.

For example, let us consider a set of eight observations of embedded boarding stops \{-2,0,3,20,-3,4,3,2\}, where each observation is a simulated boarding by a passenger where each number ($D_i$) in the set represents the difference between expected ($S_i$) and actual boarding stops ($A_i$).
With a value of 20, the fourth observation is an outlier, which might occur due to some fault in the decoder device of the public transport operator. We do not want to predict it, as it is naturally unpredictable. We seek a metric that will be both resilient to outliers, as they are unpredictable, and still account for the true dimension of the errors (see Section \ref{Ordinal classification}). Let us compare two classifiers, \textit{A} and \textit{B}.
Classifier \textit{A} predicted the following boarding stops \{-2,0,4,3,-2,3,2,2\}, while Classifier \textit{B} predicted \{3,0,3,7,1,1,3,2\}. Classifier \textit{A} is a more useful classifier since, in general, its predicted values are closer to the actual values, i.e., its variance is very small, which makes it more reliable. However, when using the classical accuracy and RMSE metrics, Classifier B has a higher accuracy and RMSE values than Classifier \textit{A}, with accuracy values of 50\% vs. 37.5\%, and RMSE values of 5.2 vs. 6. By using the Pareto Accuracy ($PA_1$), we obtain a more accurate picture in which Classifier \textit{A} clearly outperforms Classifier \textit{B} (87.5\% vs. 50\%). Here, we see a case where metrics used for both classical classification (accuracy) and ordinal classification (RMSE) do not reflect the actual performance of each classifier.

In addition to the metrics, to evaluate the performance of our model and to compare it to the schedule-based model, we also performed spatial analysis by plotting heatmaps and a temporal analysis using hours and days of the week (see Section \ref{Spatial and temporal analyses}). The analysis entailed comparing boarding stops that were predicted well, i.e., at accuracies of 50\% or above. Lastly, to enrich our understanding of the nature and patterns of PT, we produced and analyzed feature importance by exploiting the  SHAP values method, considered as constituting a unified framework for interpreting predictions based on game theory~\citep{lundberg2017unified}.\footnote{ "SHAP (SHapley Additive exPlanations) is a game-theoretic approach to explain the output of any machine learning model. It connects optimal credit allocation with local explanations using the classic Shapley values from game theory and related extensions. https://github.com/slundberg/shap"} The values are the average of the marginal contributions across all permutations.

\subsection{Procedure}\label{Procedure}
We evaluated the above methodology by applying it to the smart card data of the city of Beer Sheva, Israel. With about 200,000 inhabitants, Beer Sheva is the largest city in the southern part of Israel. It presents an interesting use case given its relatively remote location, making it more isolated from a traffic perspective. Additionally, it has a sparse PT network that is easier to model. Furthermore, it has complete passenger boarding stop information, and road traffic in the city is not prone to heavy congestion. We utilized a smart card dataset consisting of over 1M records (after preprocessing, about 92\% of the smart card records remained)  from over 85,000 distinct travelers for one month during November and December 2018. 


Next, we used a GTFS feed containing over 27,000 stops and over 200,000 PT trips in Israel for the equivalent period as the smart card data and included all the operators (or agencies in GTFS tables) in the country. The dataset also included a detailed timetable for every PT trip. Lines and stops for the city of Beer Sheva were sorted by operator and geographic coordinates. All selected routes were bus lines.

We also used a geospatial dataset from the municipal open GIS portal that contained a variety of geographical attributes of the city of Beer Sheva, such as traffic light locations, built-area densities, and more. We then extracted the 15 features from the above datasets. We converted the boarding stops from their Beer Sheva identifiers to numerical values (i.e., embedding). Lastly, we estimated an ML algorithm to classify the boarding stops and evaluated the classifier’s performance as described earlier.

\subsection{Model validation, comparative imputation methods and robustness}\label{Model validation, comparative imputation methods and robustness}
As mentioned, one of our primary goals was to develop a generic model that can be applied in any city. To that end, we validated our model based on the data of a neighboring city, Kiryat Gat situated 43Km north of Beer Sheva. We applied the method of transfer learning \citep{torrey2010transfer}, entailing the transfer of relevant knowledge by fine-tuning a model on a "novel" dataset, i.e., a set of data on which it did not train. Other than allowing our model to train more to prove our hypothesis, we split the data initially into intervals of 10 days for the transfer learning task and then into intervals of 20 days for the evaluation. 
While this grouping of the data could cause sub-par model performance, it showed that transfer learning could be accomplished with little data, most of which the model can impute. Spatial and temporal analyses were also included for this use case.

The main advantage of our modeling approach is that no ground truth is necessary to apply the model. This advantage is related to the fact that training is enabled without using domain-specific labels i.e., when data integrity is poor, and no complementary data is available. We test this assertion by comparing the ML model to other possible imputation models. Such methods, specifically passenger history and temporal closeness, can, in some cases, provide very accurate predictions, mainly when data integrity is high. However, it is important to note that they have some essential limitations. The passenger history method requires passengers have multiple observations in the dataset, which is not always available when dealing with irregular travelers or to split the research data and utilize fewer data records. Additionally, the temporal closeness method is susceptible to data integrity and sparse rides. 
Passenger history and temporal closeness were applied using the following two algorithms:

\begin{algorithm}[H] 
\caption{Predicting boarding stop based on passenger history}
\begin{algorithmic}[1]
\For{each observation $i$ $\in$ $S$}   
 \State {
IF $\exists$ $H_{i,r,t}$ return  $H_{i,b,r,t}$
   
Else  return $P_{i}$ 
}
\EndFor
\end{algorithmic}
\end{algorithm}

\begin{algorithm}[H] 
\caption{Predicting boarding stop based on temporal closeness}
\begin{algorithmic}[1]
\For{each observation $i$ $\in$ $S$}   
 \State {
IF $\exists$ $j$ |$T_j$ $-$ $T_i$| $<$ 30 return $B_j$
   
Else  return $P_{i}$ 
}
\EndFor
\end{algorithmic}
\end{algorithm}

 In addition, we also evaluated a semi-random classifier as a lower end imputation method using the following algorithm:  
\begin{algorithm}[H] 
\caption{Predicting boarding stops based on semi-random predictions}
\begin{algorithmic}[1]

\For{each boarding stop $j$}   
\State {$P_j$ $\gets$ {$\frac{\sum_{i\ =\ 1}^{n}{B_i\ =\ j}}{n}$}}
\EndFor

\For{each observation $i$ $\in$ $S$}   
 \State {
Sample $B_j$

Return $B_j$
}
\EndFor
\end{algorithmic}
\end{algorithm}

Where:
\begin{enumerate}
\item $S$ - Smart Card dataset
\item $H_{i,r,t}$ - is the history of passenger $i$ in route $r$ and time period $t$
\item $H_{i,b,r,t}$ - is the most frequent boarding stop $b$ of passenger $i$ in route $r$ and time period $t$
\item $P_{i}$ - is the ML prediction for observation $i$ 
\item $T_j$ - is the timestamp of observation $j$
\item $B_j$ - is the boarding stop of observation $j$
\end{enumerate}

Model robustness was validated by examining model performance on irregular passengers in comparison to the comparative imputation methods, given that simple imputation methods are ineffective when considering irregular travelers \citep{van2005accurate}. Therefore, we examined model performance for predicting the boarding stop of one-time travelers in Beer Sheva, i.e., passengers who boarded once and did not return with PT on the same day. These observations are usually discarded because they do not contribute to OD estimation.

\section{Results}\label{Results}
The results are presented in the following order: First, we describe some properties of the data we used, showing its suitability for the developed methodology. Second, we describe the estimated ML model and its performance in comparison to the schedule-based model. Third, we analyze the performance between the two models both temporally and spatially. Fourth, we show the validation of the ML model on the use case of the city of Kiryat Gat, using transfer learning. Fifth, we compare our model to other alternative imputation model specifications mentioned earlier. Lastly, we examine prediction robustness.

\subsection{Data properties}\label{Data properties}
\begin{figure}[H]
    \centering
  \includegraphics[scale=0.5]{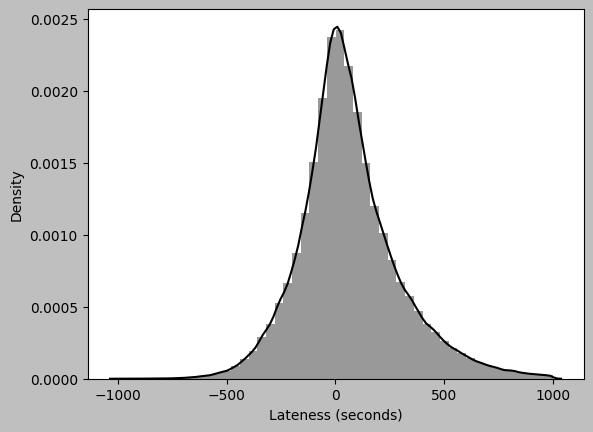}
  \caption{The density of lateness in seconds in Beer Sheva.}
  \label{fig:Lateness}
\end{figure}

We began the analyses by exploring the processed data. First, we examined the degree of lateness in the smart card data compared to the timetable data in the GTFS feed for the city of Beer Sheva. For every PT trip, the time difference between planned and actual arrival times was computed for every stop on each line (see Figure \ref{fig:Lateness}). As can be observed in Figure \ref{fig:Lateness}, the density function shows both incidents of early arrival and lateness between about 500 seconds (8 minutes) early to 1000 seconds (16 minutes) late. This result suggests that the data is very suitable for applying our method. Moreover, it can be estimated that the schedule-based model using only GTFS timetable data will be less accurate.

\begin{figure}[H]
    \centering
  \includegraphics[scale=0.7]{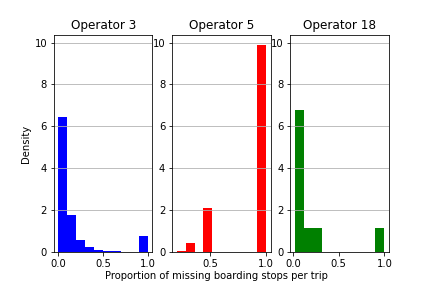}
  \caption{The Ratio of missing boarding stops per operator. Operator 3 is the largest, and 5, 18 are the second and third largest PT operators.}
  \label{fig:Null_per_operator}
\end{figure}

Second, we investigated the distribution of the missing boarding stop information in the smart card data. Figure \ref{fig:Null_per_operator} presents the mean proportion of missing boarding stops per trip of the top three PT operators in Israel. This distribution is not random. If boarding stops were missing at random, the mean would be expected to be around 0 with a long tail. However, as the density function is far from that shape, we can deduce that boarding stops are indeed not missing at random.

\subsection{Model training and performance}\label{Model training and performance}
We trained several classifiers and evaluated their performances. Among the trained classifiers, the XGBoost classifier presented the best performance  (see Table \ref{tab:full_resultse}). We compared the classifiers using the common metrics as described before. Additionally, we evaluated our Pareto Accuracy metric based on error sizes of 1, 2, i.e., $PA_1$, $PA_2$. Any larger gap would typically be deemed unacceptable in terms of level-of-service and because these error sizes are highly correlated with $PA_i$ for $i>2$. One significant advantage of embedding is the calculation speed, which was an average of $15.9\pm0.023$ seconds on about 300K observations. 

\begin{table}[H]
\caption{\label{tab:full_resultse}{Classifier Performances (Test)}}
\centering
\begin{tabular}{|l|c|c|c|c|c|c|c|}
\hline
\textbf{Algorithm}           & \textbf{Accuracy}       & \textbf{Recall}         & \textbf{Precision}      & \textbf{F1}             & \textbf{AUC}            & $\bm{PA_1}$      & $\bm{PA_2}$      \\ \hline
Schedule based      & 0.209          & 0.209          & 0.212          & 0.209          & 0.590          & 0.470          & 0.643          \\ \hline
Logistic Regression & 0.205          & 0.205          & 0.097          & 0.102          & 0.573          & 0.474          & 0.654          \\ \hline
Random Forest       & 0.368          & 0.368          & 0.348          & 0.353          & 0.666       & 0.672          & 0.818          \\ \hline
XGBoost                 & \textbf{0.410} & \textbf{0.410} & \textbf{0.393} & \textbf{0.394} & \textbf{0.765} & \textbf{0.712} & \textbf{0.843} \\ \hline
\end{tabular}
\end{table}

\begin{figure}[H]
    \centering
  \includegraphics[scale=0.4]{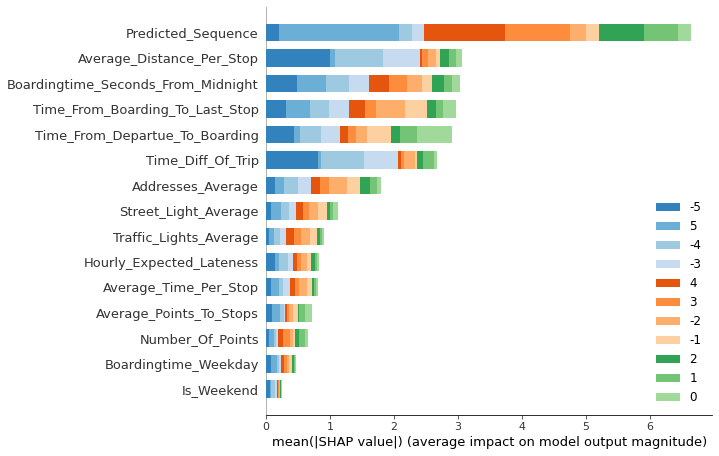}
  \caption{Feature importance using SHAP values}
  \label{fig:Feature_importance}
\end{figure}

The SHAP values to evaluate the effect of each feature are presented in  Figure \ref{fig:Feature_importance}. Here we can note: (a) by far the most important feature for the prediction is created by the predicted sequence, which shows it is highly correlated to actual patterns and is very useful for classification (i.e., schedule-based); (b) other than the first two SHAP features, the following four are temporal, which is commonsensical given that the different periods have varied impacts on traffic (such as the morning peak) and as a bus progresses along its route, stochastic events accumulate and the variance increases; (c) although geospatial features are not of the highest importance, they are not trivial, and thus, we conclude that certain physical attributes can influence the nature of our problem, e.g., denser areas can engender more congestion; and (d) the two least significant features pertain to day of the week, from which we can assert that daily PT routines remained quite stable in our case study.

\begin{figure}[h]
    \centering
  \includegraphics[scale=0.5]{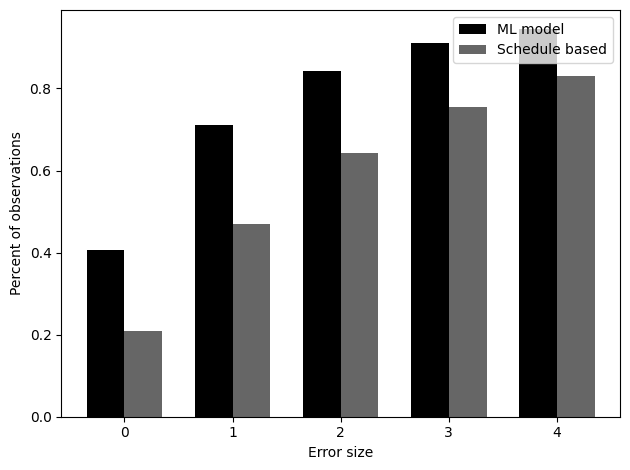}
  \caption{Pareto Accuracy comparison between ML and schedule-based models (test)}
  \label{fig:Pareto chart}
\end{figure}

In Figure \ref{fig:Pareto chart}, we  present \emph{Pareto Accuracy} between the ML model and the schedule-based one. It shows that the results are stable even for higher values than 1. Therefore, we can conclude that the proposed model outperforms the schedule-based model. 

\subsection{Spatial and temporal analyses}\label{Spatial and temporal analyses}
\begin{figure}[H]
\begin{subfigure}{.5\textwidth}
  \centering
  \includegraphics[height=5cm]{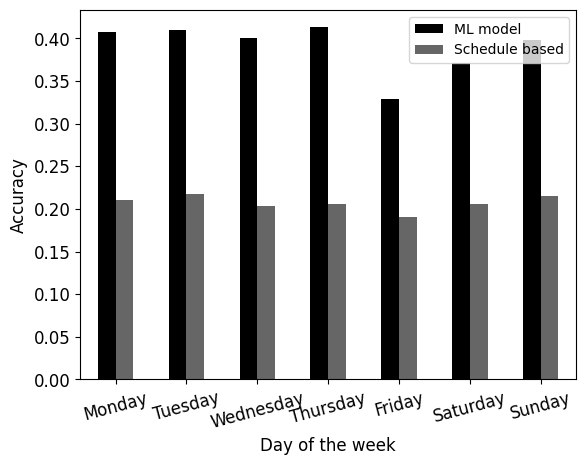}  
  \caption{Daily performance}
\end{subfigure}
\begin{subfigure}{.5\textwidth}
  \centering
  \includegraphics[height=5cm]{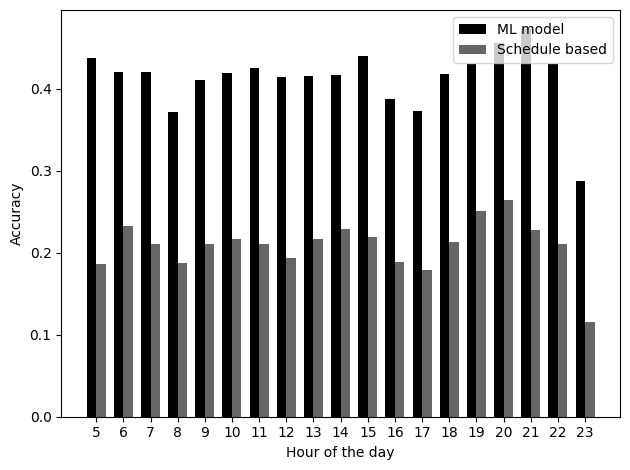}  
  \caption{Hourly performance}
\end{subfigure}
  \caption{Temporal performance of models (test) - (a) Daily, (b) Hourly}
\label{fig:Temporal performance comparison}
\end{figure}

\begin{figure}[H]
\begin{subfigure}{.5\textwidth}
  \centering
  \includegraphics[height=5cm]{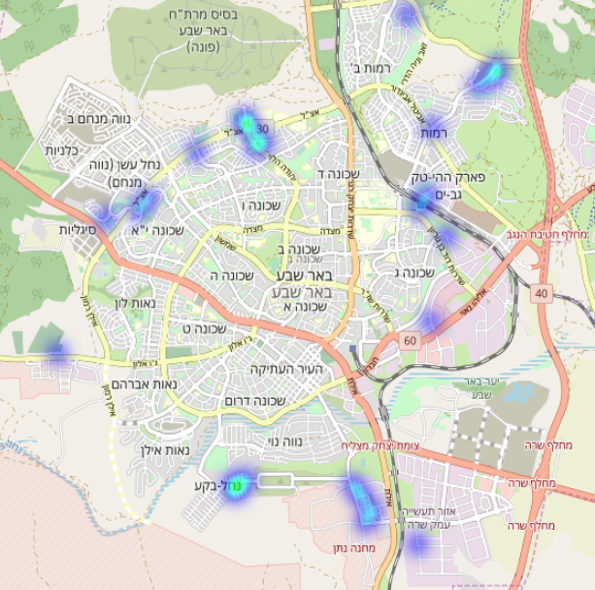}  
  \caption{Schedule-based predictions}
\end{subfigure}
\begin{subfigure}{.5\textwidth}
  \centering
  \includegraphics[height=5cm]{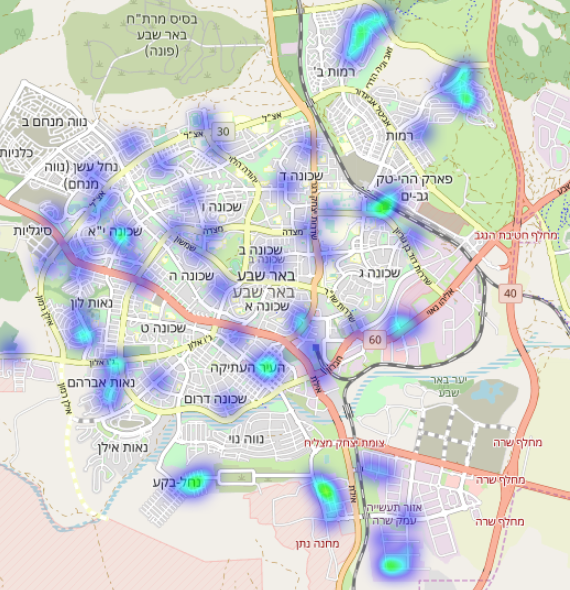}  
  \caption{Proposed model predictions }
\end{subfigure}
  \caption{Heatmaps of boarding stops with prediction accuracy of over 50\% (test)}
\label{fig:heatmaps}
\end{figure}

In addition to the aggregated results, we analyzed the model performance both temporally  (see Figure \ref{fig:Temporal performance comparison}) and spatially (see Figure \ref{fig:heatmaps}). The temporal analysis shows that, in terms of accuracy,  our proposed model outperformed the schedule-based method on both a daily and an hourly basis.\footnote{the hourly analysis was done for weekdays when the presence of traffic congestion makes the prediction of PT service punctuality likely more difficult}.
Moreover, the spatial analysis showed similar results, and the stops where the predictions were ranked 'good', i.e., over 50\% accuracy, were plotted.

Two major insights can be derived from these analyses:
First, the ML model predicts many more stops than the schedule-based model. Second, the schedule-based model renders good predictions mainly for the central stops (train stations, main roads, or industrial zones). However, when the model is applied to non-central locations, it is suboptimal, in stark contrast to the ML model, making good predictions across all locations.

\subsection{Model validation }\label{Model validation }
\begin{figure}[h]
    \centering
  \includegraphics[scale=0.5]{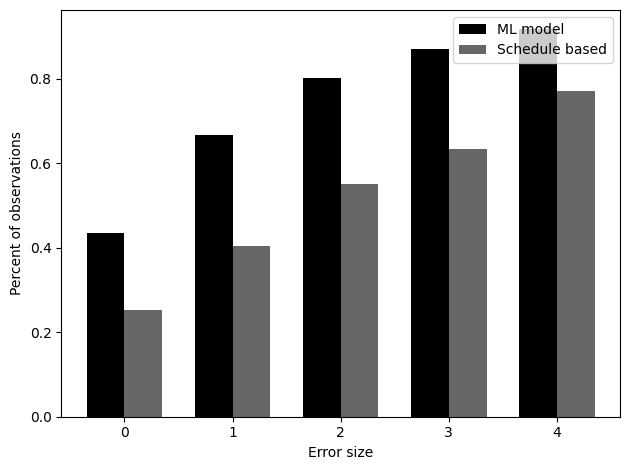}
  \caption{Pareto Accuracy comparison of models between ML and schedule-based models (validation)}
  \label{fig:Pareto chart Kiryat Gat}
\end{figure}

\begin{figure}[H]
\begin{subfigure}{.5\textwidth}
  \centering
  \includegraphics[height=5cm]{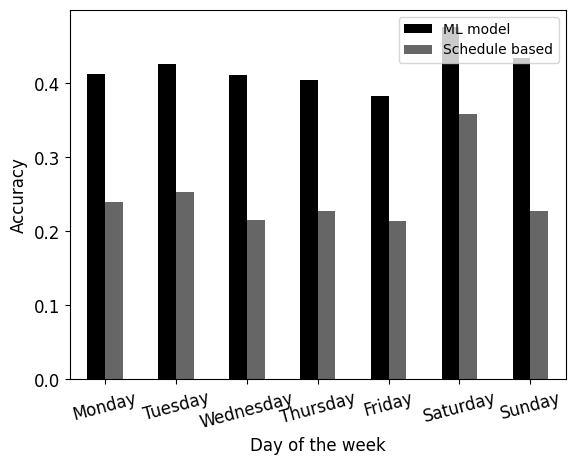}  
  \caption{Daily performance comparison.}
\end{subfigure}
\begin{subfigure}{.5\textwidth}
  \centering
  \includegraphics[height=5cm]{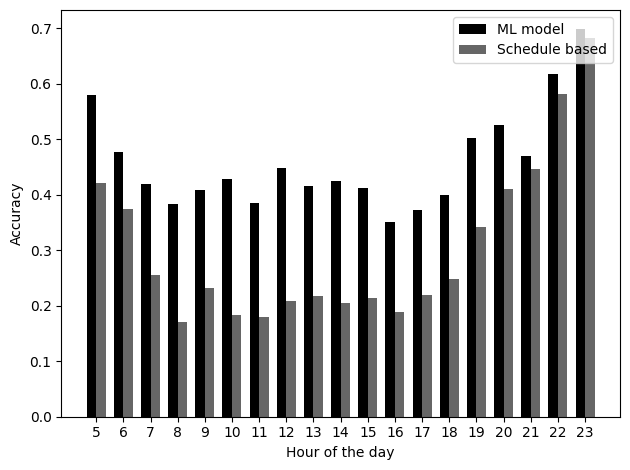}  
  \caption{Hourly performance comparison.}
\end{subfigure}
  \caption{Temporal performance of models (validation).}
\label{fig:Temporal performance comparison Kiryat Gat}
\end{figure}

\begin{figure}[H]
\begin{subfigure}{.5\textwidth}
  \centering
  \includegraphics[height=5cm]{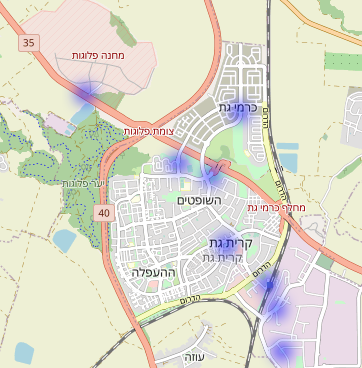}  
  \caption{Schedule-based predictions.}
\end{subfigure}
\begin{subfigure}{.5\textwidth}
  \centering
  \includegraphics[height=5cm]{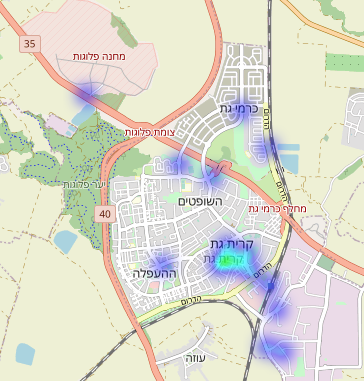}  
  \caption{Proposed model predictions.}
\end{subfigure}
  \caption{Heatmaps of boarding stops with predicted accuracy of over 50\% (validation).}
\label{fig:heatmaps_kiryat_gat}
\end{figure}

As noted, we performed the model validation for the nearby city of Kiryat Gat. Evidently the ML model performed remarkably better than the schedule-based model (see Table \ref{tab:full_resultse_kiryat_gat}). Figure \ref{fig:Pareto chart Kiryat Gat} shows the Pareto Accuracy for different values of error size showing the ML model is consistently better. \footnote{We only used the XGBoost model as it was the best performing model on the test dataset.} Figure \ref{fig:Temporal performance comparison Kiryat Gat} presents performance comparison in Kiryat Gat shows the temporal analysis – accuracy by day of the week and on an hourly basis for weekdays which shows similar properties to the trained model the ML model demonstrated higher accuracy compared to the schedule-based model. Figure \ref{fig:heatmaps_kiryat_gat} presents the spatial analysis revealing once more that the ML model predicts more stops with higher accuracy.

\begin{table}[H]
\caption{\label{tab:full_resultse_kiryat_gat}{Classifier performances for model validation}}
\centering
\begin{tabular}{|l|c|c|c|c|c|c|c|}
\hline
\textbf{Algorithm} & \textbf{Accuracy} & \textbf{Recall} & \textbf{Precision} & \textbf{F1}    & \textbf{AUC}   & \textbf{$\bm{PA_1}$}   & \textbf{$\bm{PA_2}$}   \\ \hline
Schedule based     & 0.253             & 0.253           & 0.224              & 0.234          & 0.599 & 0.404          & 0.550          \\ \hline
XGBoost            & \textbf{0.438}    & \textbf{0.438}  & \textbf{0.441}     & \textbf{0.419} & \textbf{0.685}        & \textbf{0.668} & \textbf{0.802} \\ \hline
\end{tabular}
\end{table}

\subsection{Comparative imputation methods}\label{Comparison to other imputation models}
Table \ref{tab:Other methods} shows the results of the comparisons to alternative imputation methods. While the predicted accuracy of the two alternative methods is similar, the  disadvantages of the aforementioned methods are more evident in the lower share of the population than can be predicted compared to the ML model. The semi-random classifier naturally demonstrates that it is far from trustworthy in the case of hierarchical PT networks.

\begin{table}[H]
\caption{\label{tab:Other methods}Results of comparative imputation methods}
\centering
\begin{tabular}{|l|c|c|}
\hline
\textbf{Method} &
  \multicolumn{1}{l|}{\textbf{Percent predicted}} &
  \multicolumn{1}{l|}{\textbf{\begin{tabular}[c]{@{}l@{}}Accuracy for \\ predicted \\ observations\end{tabular}}} \\ \hline
Proposed XGBoost                                                       & 100\% & 41\% \\ \hline
Passenger history                                                      & 82\%  & 59\% \\ \hline
\begin{tabular}[c]{@{}l@{}}Temporally close \\ passengers\end{tabular} & 52\%  & 59\% \\ \hline
\begin{tabular}[c]{@{}l@{}}Semi-random\\ guessing\end{tabular}         & 100\% & 11\% \\ \hline
\end{tabular}
\end{table}

It is important to note that while the accuracy of our proposed method is lower, it is far more robust, both in terms of percentage of population predicted and on irregular travelers, which other suggested methods are incapable of predicting (see Section \ref{Prediction robustness to irregular travelers}). For example, in predicting using historical records, we cannot predict for a new passenger or a new route. For using temporal closeness, the prediction will be extremely sensitive to sparse routes.

\subsection{Robustness to irregular travelers}\label{Prediction robustness to irregular travelers}

While the personal history method can indeed be relevant as evident in Table \ref{tab:Other methods}, as noted above (see Section \ref{Model validation, comparative imputation methods and robustness}) model robustness was evaluated examining performance for predicting the boarding stop of one-time travelers. As shown in Table \ref{tab:Results_on_iregular_travelers}, the results clearly show (see first row in Table \ref{tab:Results_on_iregular_travelers}) that the ML model is robust and capable of predicting missing stops even for irregular or new passengers that have no historical pattern. Additionally, as noted earlier, the suggested methods are very limited. The evaluation of passengers they do not predict is clearly shown below in Table \ref{tab:Results_on_iregular_travelers} (see second and third row).

\begin{table}[H]
\caption{\label{tab:Results_on_iregular_travelers}{Results of ML model (XGBoost) on one time traveler and passengers not predicted by comparative methods suggested in Table \ref{tab:Results_on_iregular_travelers}}}
\centering
\begin{tabular}{|l|c|c|c|c|c|c|c|}
\hline
\textbf{Passenger type} &
  \textbf{Accuracy} &
  \textbf{Recall} &
  \textbf{Precision} &
  \textbf{F1} &
  \textbf{AUC} &
  \textbf{$\bm{PA_1}$} &
  \textbf{$\bm{PA_2}$} \\ \hline
\begin{tabular}[c]{@{}l@{}}One time\end{tabular} &
  0.408 &
  0.408 &
  0.394 &
  0.390 &
  0.767 &
  0.703 &
  0.838 \\ \hline
\begin{tabular}[c]{@{}l@{}}Not predicted \\ by method 1\end{tabular} &
  0.419 &
  0.419 &
  0.407 &
  0.402 &
  0.772 &
  0.706 &
  0.835 \\ \hline
\begin{tabular}[c]{@{}l@{}}Not predicted \\ by method 2\end{tabular} &
  0.348 &
  0.348 &
  0.336 &
  0.330 &
  0.744 &
  0.671 &
  0.822 \\ \hline
\end{tabular}
\end{table}

\section{Discussion and conclusions}\label{Disccusion}
In this study, we showed that by mining smart card data and extracting timetable data, we could construct a passenger boarding stop prediction model, which surpasses the traditional schedule-based method. Our research revealed that applying machine learning techniques improves the integrity of PT data, which can significantly benefit the field of transportation planning and operations. From the results, we can deduce the following conclusions: First, our methodology for feature extraction and machine learning model construction demonstrates several noteworthy advantages: (a) the ML algorithm generates a \textit{generic model} that can be used with other smart card datasets since the labels (i.e., numeric representations) are always aligned in all datasets; (b) by embedding the boarding stops, our method ensures that the number of distinct labels is relatively small and significant \textit{computation time reduction} can be accomplished; (c) boarding stop use is inherently imbalanced, as some stops are frequently used while others are used rarely. Our proposed methodology is able to accurately classify many classes despite the inherent imbalances, thus contributing to \textit{unpredictability reduction}; (d) the method is \textit{data lean} and requires only mining a smart card dataset and a GTFS feed (or any compatible timetable dataset) without the need to process any other datasets; (e) the ML model is entirely \textit{complementary} to other imputation methods including the schedule-based method as well as passenger history or temporal closeness; and (f) the method \textit{provides a robust model} capable of dealing even with irregular or unpredictable passengers.

Second, our model (applying the XGBoost algorithm) produced the highest performance, with 41\% accuracy and 71\% $PA_1$, whereas the schedule-based method achieved only 21\% accuracy and 47\% $PA_1$. Even for larger error sizes, the ML model outperformed the schedule-based one. Moreover, the schedule-based method was able to render good predictions only for a few main stops compared to the ML model, which predicted well across all stops. This dependency on centrality was clearly visible in the spatial analysis of the stops that were well-predicted. This result confirms our conjecture that the schedule-based imputation approaches can be significantly improved by using ML methods. Furthermore, we also found that complex methods, such as ensemble, resulted in much better model performance than simple algorithms, such as logistic regression. In future research, we intend to test the performance of additional prediction algorithms, such as Deep Neural Networks \citep{jung2017deep, liu2017novel}.

Third, from the SHAP values (Figure \ref{fig:Feature_importance}), the following can be noted: The temporal features (created by the timetable from the GTFS feeds) are indeed crucial for the operation of the ML model. Geospatial features, however, were less important. Accordingly, we estimated a model trained without the geospatial features (see Table B.1 in Appendix B). In comparison to the richer model, the performance is somewhat worse. Therefore, we assert that such information is considered useful: Firstly, to understand patterns in a given city, for instance, which spatial attribute is more closely correlated with lateness or earliness. Secondly, it can help the transfer learning process in a new city, i.e., if the model was trained on city A, and will be used to predict city B, using the spatial features will produce a more robust model to the difference between those cities.

Fourth, we showed that the  ML model is transferable (see Section \ref{Feature extraction and machine learning model construction}) and able to provide strong and consistent results when validated on another city while outperforming the schedule-based imputation method. Nonetheless, our method, given its generic nature, is not entirely comparable with methods of dissimilar nature, such as those presented in Table \ref{tab:Other methods} which cannot be straightforwardly transferred to another context. Since, to the best of our knowledge, no other imputation method shows such transferability, robustness, and generic nature other than the schedule-based imputation, the latter should be regarded as the comparative benchmark until another imputation method is developed. 

Fifth, we recommend using our model when the lack of data does not allow for other more accurate methods to be used, such as passenger history or temporal closeness. Nonetheless, our model can complement these methods, especially for those records that are overlooked, as shown in Table \ref{tab:Other methods}, and thus can utilize more of the scarce data at hand. As noted, our method does not require to mine or access any additional datasets (like AVL or APC), which are not always available and can increase the extent of errors in the prediction. This observation makes our method extremely suitable for planning purposes in non-auto-dependent and less technologically-orientated societies in the developing countries and the Global South \citep{sohail2006effective}.

Lastly, we introduced a new generalized accuracy metric which we named Pareto Accuracy that allows to better compare between classifiers for ordinal classification problems. This metric is more robust to outliers, easier to interpret, and accounts for the true dimension of errors. In addition, the metric is easy to implement. In the future, we hope to understand how Pareto Accuracy can improve additional ordinal classification use cases.

There are a few limitations to the study worth noting. One is that our method requires several constraints to succeed, such as: timestamps, trip IDs, and existing trip timetables. These constraints potentially reduce the number of relevant datasets and the number of observations that could be imputed. However, these constraints also preclude the use of the schedule-based method; hence, in practice, our method has little effect on the ability to impute missing data. In addition, the generality of our method can increase bias, as it ignores features that cannot be transferred between datasets. These features, such as having each PT line as a categorical feature, can reduce bias when imputing a specific dataset.

Possible extensions include: predicting alighting stops (when the operator does not record TAP out), imputing other attributes of interest such as trip ID or time of day, etc. In the future, we would like to test our model in other cities to verify its generalizations. In addition, we also suggest testing the influence of  transfer learning on new datasets. 

To summarize, missing data imputation is a difficult and complex task. On the one hand, one wants as much data as possible for analyses, while on the other hand, data integrity is of critical importance and demands the availability of imputation methods that work well. We assert that the commonly used schedule-based method suffers from a subpar performance in terms of accuracy and other key metrics. It is highly dependent on the centrality of boarding stops. In contrast, we showed that our model outperformed the schedule-based method in all metrics over different temporal periods. It was more robust to the centrality of the imputed stops and irregularity of recorded trips. This makes it a much more suitable method for imputation as it improves data integrity. In addition, our method is based on generic classification and thus can be used in a wide variety of use cases.

\section*{Acknowledgements}
This research was supported by the Ministry of Science \& Technology, Israel, and The Ministry of Science \& Technology of the People's Republic of China (Grant No. 3-15741). We want to thank Prof. Itzhak Benenson (Tel Aviv University) for preliminary discussions of the research question. We also want to thank Valfredo Macedo Veiga Junior (Valf) for designing the infographic illustration and Sandra Falkenstein for editing and proofreading this paper. Special thanks to Data Scientist Raz Vais, advisor to the Israeli National Public Transport Authority, for help in obtaining and processing the data.

\bibliographystyle{apalike}  
\bibliography{Paper}

\section*{Appendix A}
Metrics presented in this paper:
 \begin{itemize}
  \item \textit{Accuracy} - Percent of observations that were correctly classified
 \item \textit{Recall} - The number of observations for each class that were correctly classified divided by the total number of distinct observations from this class. Final \textit{Recall} is the weighted average of the above on all classes. 
 \item \textit{Precision} - The number of observations for each class that were correctly classified divided by the total number of observations that were predicted within this class. Final \textit{Precision} is the weighted average of the above on all classes. 
  \item \textit{F1} - 2*(Precision * Recall)/(Precision + Recall)
  \item \textit{AUC} - Area under curve (AUC) is the area under the ROC curve. This curve, for each class, is the true positive rate as a function of the false positive rate. A weighted average of the areas under the curves of all classes is calculated as the AUC metric.
   \item \textit{RMSE} - Root mean square error (RMSE) is a method for ordinal classification and regression. It sums the square difference from prediction to actual label, then returns the root of the above average.
\end{itemize}

\section*{Appendix B}
\begin{table}[H]
\caption{Model performance with and without geospatial features (XGBoost)}
\centering
\begin{tabular}{|l|c|c|c|c|c|c|c|}
\hline
\textbf{Algorithm} & \textbf{Accuracy} & \textbf{Recall} & \textbf{Precision} & \textbf{F1} & \textbf{AUC} & \textbf{$\bm{PA_1}$} & \textbf{$\bm{PA_2}$} \\ \hline
Without            & 0.409             & 0.409           & 0.392              & 0.393       & 0.770        & 0.712        & 0.842        \\ \hline
With               & 0.410             & 0.410           & 0.393              & 0.394       & 0.765        & 0.712        & 0.843        \\ \hline
\end{tabular}
\end{table}

\end{document}